\documentclass[runningheads]{llncs}
\usepackage[T1]{fontenc}
\usepackage{graphicx}

\usepackage{amsmath} 
\usepackage{times}  
\usepackage{helvet}  
\usepackage{courier}  
\usepackage[hyphens]{url}  
\usepackage[numbers,square]{natbib}
\usepackage{caption} 
\usepackage{algorithm}
\usepackage{algorithmic}
\usepackage{amssymb}
\usepackage{booktabs}
\usepackage{newfloat}
\usepackage{listings}
\usepackage{placeins}
\usepackage{marvosym}

\begin{document}

\title{A hierarchy tree data structure for behavior-based user segment representation}

\author{
Yang Liu\inst{1} \orcidID{0000-0001-7490-8597} ~(\Letter),
    Xuejiao Kang\inst{1}, 
    Sathya Iyer\inst{1}, 
    Idris Malik\inst{1} \orcidID{0009-0008-2443-8919}, 
    Ruixuan Li\inst{1} \orcidID{0009-0008-3384-2251}, 
    Juan Wang\inst{1} \orcidID{0009-0000-1665-0567},
    Xinchen Lu\inst{1},
    Xiangxue Zhao\inst{1},
    Dayong Wang\inst{1},
    Menghan Liu\inst{1},
    Isaac Liu\inst{1},
    Feng Liang\inst{1},
    Yinzhe Yu\inst{1} 
}
\authorrunning{Y. Liu et al.}

\institute{
Meta, Menlo Park, California, USA 94025 
    \email{
 \{yliu9,xuejiao,sath,idrismalik,ruixuan,juanw,xinchenlu,sherryz, 
 \\dayongwang,menghanliu,iliu,liangfeng,yinzheyu\}@meta.com
    }
}
\maketitle  

\begin{abstract}
   User attributes are critical in modern recommendation systems, especially for alleviating cold-start challenges and enhancing experiences for new or infrequent users. Integrating diverse categorical attributes, such as demographics and interests, in a scalable and behavior-aware manner remains challenging. We present Behavior-based User Segmentation (\textit{BUS}), a tree-based framework that hierarchically partitions users guided by product-specific engagement behaviors. During tree construction, a greedy algorithm optimizes Normalized Discounted Cumulative Gain (NDCG) to ensure behavioral representativeness between marginal and active users. A novel $regress$ operator filters and aggregates irrelevant user attributes using loss signals at each iteration. The \textit{BUS} tree enables aggregation across leaf and internal nodes to identify representative user segments and generate popular content and interaction patterns. To improve fairness and robustness, \textit{BUS} incorporates connection-based segmentation via social graphs, combining individual and social behavioral patterns. Deployed at industrial scale, \textit{BUS} serves billions of users daily, achieving significant improvements in metrics such as music ranking and email notifications delivery time. To our knowledge, \textit{BUS} is the first list-wise learning-to-rank framework for tree-based recommendation that integrates diverse categorical attributes with semantic interpretability at industrial scale.

\end{abstract}

\section{Introduction}

User attributes provide structured representations of user-related signals that guide personalization in modern recommendation systems. These attributes include relatively static demographic and technical features (e.g., age, gender, device type) as well as dynamic behavioral and interest-based representations (e.g., preferred content categories or long-term engagement topics). Together, they capture both stable and evolving aspects of user profiles, offering a rich context for modeling preferences and improving personalization \cite{adomavicius2005toward}.

Demographic-based recommendation, one of the earliest approaches, assumes that users with similar demographic characteristics tend to have similar preferences \cite{pazzani1999framework}. This method is transparent, interpretable, and effective for new or infrequent users, addressing the cold-start problem and enabling easily explainable recommendations compared to complex deep learning-based models \cite{al2016user}.

\begin{figure}[H]
\begin{minipage}{1.0\linewidth}
  \includegraphics[width=\linewidth]{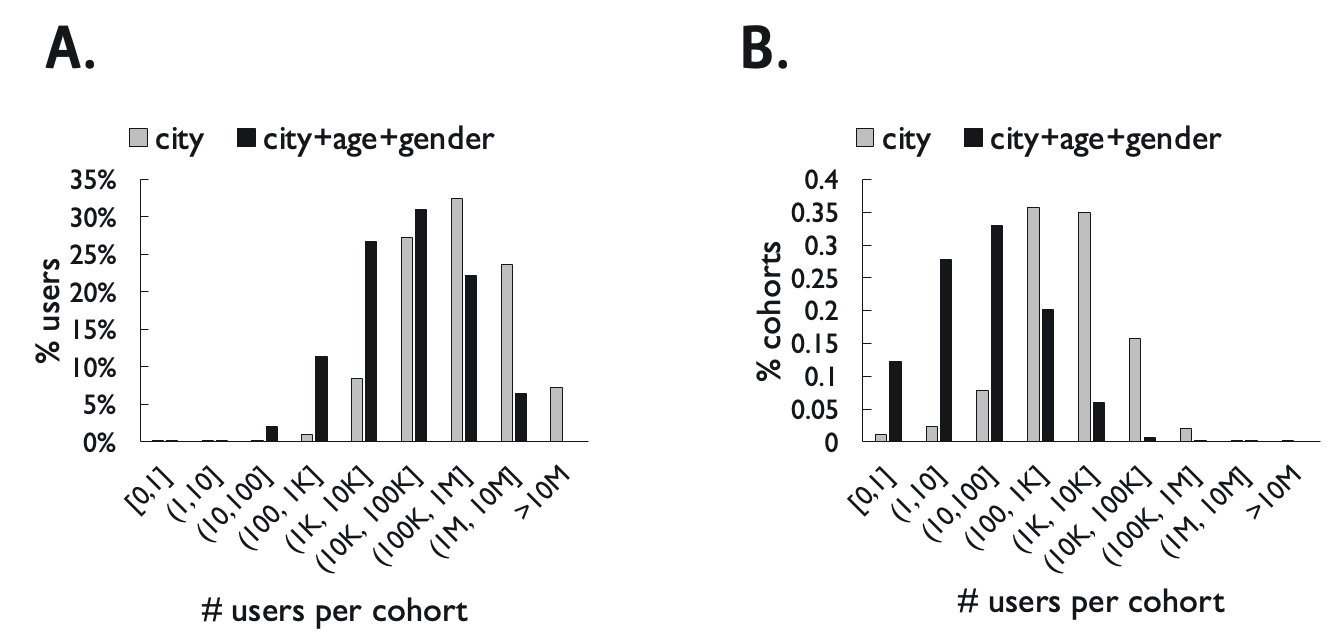}
  \captionof{figure}{User cohorts size. (A) User distribution. (B) Cohort distribution}
  \label{fig:cohort}
\end{minipage}
\end{figure}

However, attribute-based recommendation faces several limitations. First, it often provides limited personalization and may amplify bias, leading to repetitive and homogeneous recommendations \cite{fayyaz2020recommendation}. Second, it largely depends on empirically derived user attributes that are specific to certain product contexts, making them difficult to generalize across products with distinct market positions or objectives. Third, categorical features often exhibit high cardinality and dependency, posing challenges for efficient processing, fairness, and scalability. Scaling user attribute-based recommendation to support diverse products and billions of users poses substantial challenges in both complexity and scalability. The large and heterogeneous user base of each product requires highly efficient algorithms capable of processing massive volumes of behavioral and attribute data. When multiple attributes are directly combined or aggregated (Figure~\ref{fig:cohort}), controlling user segment size becomes difficult, and skewed demographic distributions can lead to biased or unrepresentative segments. Furthermore, the dynamic nature of user behavior—with evolving demographic profiles and shifting preferences—necessitates adaptive learning mechanisms that continuously update attribute-based models to capture the latest user - product engagement pattern. Notably, deep learning approaches such as neural collaborative filtering and graph neural networks require sufficient interaction history to learn meaningful representations, leaving cold-start and marginal users underserved.

\setlength{\abovecaptionskip}{2pt}
\setlength{\belowcaptionskip}{2pt}
\renewcommand{\arraystretch}{0.9}
\vspace{-2em}
\vspace{-2em}

In this paper, we reformulate user attribute-based recommendation as a listwise learning-to-rank problem, leveraging diverse categorical features to construct targeted user segments that transfer knowledge from active users to new or infrequent (marginal) users. To address this, we propose a novel tree-based data structure, \textit{BUS}, as the core of our approach. We describe the construction of the \textit{BUS} tree, including attribute selection and aggregation through the $regress$ operator. Additionally, we extend \textit{BUS} with connection-based segments derived from the social graph to enhance recommendation quality and ecosystem effects. Finally, we demonstrate \textit{BUS}-based recommendation in two real-world applications: candidate retrieval for music ranking and delivery timing optimization in email notifications.

Our main contributions are as follows:
\begin{itemize}

    \item We introduce a novel tree-based data structure, \textit{BUS}, to represent user attribute segments, and productionize the data structure and training algorithm in SQL.
    \item We propose a listwise learning-to-rank solution leveraging \textit{BUS} to generate product-specific segments and enable \textit{BUS}-based candidate retrieval for recommendation.
    \item We evaluate our approach on production traffic serving billions of users daily and demonstrate substantial improvements in key metrics across multiple applications.
\end{itemize}

\section{Related work}
\label{related_work}
Demographic-based user targeting and behavior modeling are widely adopted strategies in large-scale social media, search, and e-commerce platforms. Incorporating demographic attributes into recommendation systems enhances accuracy and preserves users’ local experiences \cite{gon2021local}. Beyond serving as ranking features or filters, demographic attributes can also support retrieval for cold-start users \cite{al2016user}. However, prior studies often examine only fixed orderings of attributes without a principled learning algorithm to weight their contributions. Attribute selection is frequently driven by heuristics or experimentation rather than systematic, data-driven methods.

Most demographic attributes are categorical and can be encoded via integer encoding, one-hot encoding, or entity embeddings for deep learning \cite{hancock2020survey}. While embeddings improve memory efficiency and capture intrinsic feature properties \cite{guo2016entity, zhang2022city2vec}, they reduce interpretability, complicating debugging and signal quality assessment. Recently, Large Language Models (LLMs) have been applied to convert structured user attributes into textual descriptions, which are then embedded and clustered into user groups using methods such as K-Means \cite{sun2024random, li2025consumer, tissera2024enhancing}. These approaches, however, largely ignore user–product interactions and rely on heuristic clustering without optimizing cluster granularity or downstream ranking quality at scale.

Beyond flat encodings, hierarchical and tree-based structures have been explored to 
capture richer user representations. HieRec \cite{qi2021hierec} constructs a hierarchical 
user interest tree for news recommendation, capturing fine-grained interest representations. Tree-based indexing has also been explored for efficient retrieval, with joint optimization of tree-based index structures and deep models shown to improve recommendation efficiency \cite{zhu2019joint}. The role of user segmentation granularity in recommendation performance has been examined \cite{erdem2023role}, highlighting the importance of segment design. Other approaches model user behavior through attention mechanisms \cite{zhou2018atrank} or leverage graph neural networks with metapath aggregation for heterogeneous settings \cite{fu2020magnn}. However, these methods typically require sufficient interaction history and focus on active users rather than cold-start scenarios.

Listwise learning-to-rank (LTR) methods directly optimize ranking quality by considering the entire list of items for each query as a training instance, rather than evaluating items individually or in pairs. It has been extensively studied in information retrieval and recommendation systems, often outperforming pointwise and pairwise approaches \cite{chapelle2011future, xia2008listwise}. Algorithms such as ListNet \cite{cao2007learning}, RankCosine \cite{qin2008query}, ListMLE \cite{xia2008listwise}, SQL-Rank \cite{wu2018sql}, and GFN4Rec \cite{liu2023generative} improve ranking quality but are typically limited to small candidate sets. Applying listwise approaches to large candidate pools, especially for cold-start users, remains largely unexplored. Content-aware listwise collaborative filtering addresses new items \cite{ravanifard2021content}, but effective solutions for new or infrequent users are still needed.

\section{Methodology}
\label{methodology}
\subsection{Problem overview}
We formulate user attribute-based recommendation as an optimization problem to identify an optimal set of user segments defined by categorical attributes (e.g., age, gender, location, interest). Since active users usually provide sufficient behavioral signals, our goal is to ensure that aggregated active-user behaviors can effectively represent marginal users. Let $E_{u,p}$ denote the engagement behavior of user $u$ with product $p$, represented as a ranked list of items the user has interacted with. For active users in segment $s$, we compute the aggregated top-$K$ behavior $\hat{E}_{AU_s,p}$ by ranking items based on their popularity within the segment. We define a listwise loss for each product $p$ as:
\[
L_{T_p} = -\sum_{s \in T_p} \sum_{u \in MU_s} f(E_{u,p}, \hat E_{AU_{s,p}})
\]
where $f$ is the listwise function (e.g., NDCG) to measure how well the aggregated active-user behavior $\hat{E}_{AU_s,p}$ predicts marginal user $u$'s actual engagement $E_{u,p}$. The objective is to construct a segment tree $T_p'$ that minimizes this loss:

\[
T_p' = \arg \min_{T_p} L_{T_p}.
\]

\subsection{User segment tree}
Marginal users typically exhibit limited product engagement activity and weak graph connections, making demographic attributes the primary modeling signals. However, traditional encoding methods (e.g., one-hot, label encoding) lose semantic meaning and perform poorly with high-cardinality features. To address this, we construct a hierarchical tree-based structure that preserves semantic dependencies among categorical attributes (e.g., \textit{San Francisco} city nested under \textit{US} country). Each user is mapped to a unique leaf node based on their attribute path, enabling semantically coherent segmentation such as $global \rightarrow US \rightarrow 30s \rightarrow California \rightarrow San\_Francisco$.

We initialize the \textit{BUS} tree with root node $global$ containing all users in universe $U$, and iteratively grow tree $T$ following Algorithm~\ref{algorithm:construction}. At each level $i$, we consider eligible attribute types $Attrs = {\tau_1,\tau_2,...,\tau_n}$ and generate staging child nodes for each attribute $\tau$. Hierarchical dependencies (e.g., country before city) can be enforced for semantic consistency. For each staging node $s_{i_k}$, we aggregate top-K product behaviors $\hat{E}_{AU_{s_{i_k}},p}$ from active users (e.g., top 100 clicked items), forming a predicted ranking list. The list is compared with marginal users’ actual engagement lists using normalized discounted cumulative gain (NDCG@K) \cite{jarvelin2002cumulated} to compute the node loss.

\[
L_{s_{i_k}} = \sum_{u \in MU_{s_{i_k}}}-NDCG_u@K = \sum_{u \in MU_{s_{i_k}}}-\frac {DCG_u} {IDCG_u} 
\]
We assign zero relevance when the predicted list fails to cover the user’s engaged items.

\setlength{\abovecaptionskip}{2pt}
\setlength{\belowcaptionskip}{2pt}
\renewcommand{\arraystretch}{0.9}
\vspace{-2em}
\begin{algorithm}[H]
\caption{\textit{BUS} Tree Construction}
\begin{algorithmic}[1]
\STATE Initialize attribute list $Attrs$, user universe $U = AU_{global} \cup MU_{global}$, and root node $S_{global}$.
\STATE Compute global top-$K$ behaviors $\hat{E}_{AU_{global},p}$ and initialize loss $L = \{-\text{NDCG}(E_{u,p}, \hat{E}_{AU_{global},p}) : u \in MU_{global}\}$.
\WHILE{$|Attrs| > 0$}
  \FOR{each attribute $\tau \in Attrs$}
    \FOR{each leaf node $s_{i-1}$ in $T$}
      \STATE Split $s_{i-1}$ by $\tau$ into child nodes $s_i = \{s_{i_1},...,s_{i_n}\}$.
      \FOR{each $s_{i_k}$}
        \STATE Compute top-$K$ behaviors $\hat{E}_{AU_{s_{i_k}},p}$ and loss $L_{s_{i_k}} = \sum_{u \in MU_{s_{i_k}}} -\text{NDCG}(E_{u,p}, \hat{E}_{AU_{s_{i_k}},p})$.
        
        \STATE Compute inherited loss $\hat L_{s_{i_k}}$ from parent $s_{i-1}$.
        \IF{$\sum L_{s_{i_k}} > \omega \sum \hat L_{s_{i_k}}$ or $|AU_{s_{i_k}}| < \mu$}
          \STATE Replace $i_k$ with ``\textit{regress}'' and append $\hat L_{s_{i_k}}$ to $L$.
        \ELSE
          \STATE Append $L_{s_{i_k}}$ to $L$.
        \ENDIF
      \ENDFOR
      \STATE Record total loss $L_\tau = \sum_{s \in s_i} L_s$.
    \ENDFOR
  \ENDFOR
  \STATE Select $\hat\tau = \arg\min_\tau L_\tau$, grow $T$ with its child nodes, and remove $\hat\tau$ from $Attrs$.
\ENDWHILE
\end{algorithmic}
\label{algorithm:construction}
\end{algorithm}
\vspace{-2em}

\begin{figure}[H]
\begin{minipage}{1.0\linewidth}
    \centering
    \includegraphics[width=0.9\linewidth]{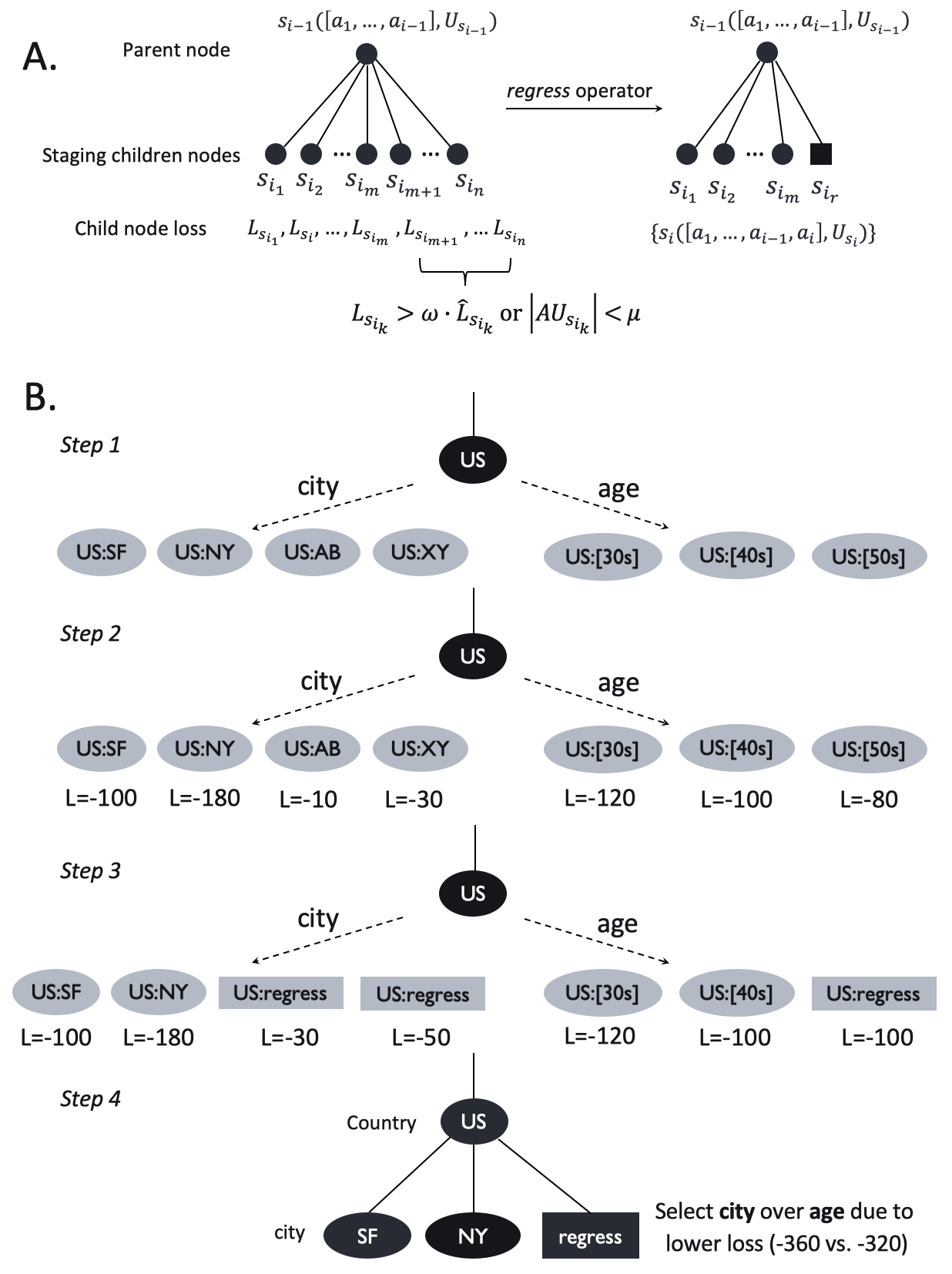}
    \caption{\textit{BUS} tree construction: (A) The \textit{regress} operator generates a regress node, shown as a square; (B) Example of user attribute selection in \textit{BUS} tree construction}
    \label{fig:construction}
\end{minipage}
\end{figure}

Because NDCG is non-differentiable, we adopt a greedy optimization strategy to ensure monotonically improved ranking quality during tree construction. Specifically, a \textit{regress} operator is introduced to compare each node’s loss $L_{s_{i_k}}$ with its inherited parent loss $\hat L_{s_{i_k}}$. A staging node is retained only if its loss is no greater than its parent’s; otherwise, it is replaced by a $regress$ node that reverts to the parent’s aggregate behavior (Figure~\ref{fig:construction}). The $regress$ tree nodes have a clear real-world interpretation. For instance, as shown in Figure~\ref{fig:construction}B, the $regress$ nodes derived from the city attribute correspond to all $US$ users not residing in San Francisco or New York; their behavior is represented using the aggregate activity patterns of all $US$ users.

\setlength{\abovecaptionskip}{2pt}
\setlength{\belowcaptionskip}{2pt}
\renewcommand{\arraystretch}{0.9}
\vspace{-2em}
\vspace{-2em}

Two hyperparameters, $\omega$ (default 1.0) and the minimum active user count $\mu$, are introduced in the training configuration to regulate tree growth. The threshold $\omega$ controls the strictness of the regress operator: lower values (e.g., $\omega < 1.0$) allow more aggressive splitting, while $\omega \geq 1.0$ ensures strictly monotonic loss reduction (Lemma 1). In our experiments, we set $\omega = 1.0$ and use the minimum active user count $\mu$ to control the segment size explicitly.

At each level, the attribute type with the minimum total loss $L_\tau$ is selected for expansion, yielding a hierarchically optimized and interpretable user segmentation iteratively.

\textbf{Lemma 1}: The overall loss during \textit{BUS} tree construction decreases monotonically when $\omega \geq 1.0$.

\textbf{Proof}: For each staging child node, two cases arise. If the node’s loss exceeds its parent’s, the \textit{regress} operator replaces it with the parent’s aggregate behavior, ensuring the child’s loss equals the parent’s. If the node’s loss is equal to or lower than the parent’s, it is retained. In both cases, each child node has loss no greater than its parent, guaranteeing monotonic decrease in overall loss.

Each user is assigned to exactly one leaf node, satisfying the MECE principle, and the resulting tree has uniform depth. The algorithm is practical to implement and scalable. Its time complexity is $O(m^2 \cdot n \cdot K \cdot C)$, where $m$ is the number of attribute types, $n$ the number of marginal users, $K$ the ranking depth used for NDCG calculation, and $C$ the cost of computing top-K behaviors. Training billions of users with 10–15 attributes typically requires 40–50 
Batch Compute Units (BCUs), far less than deep learning embedding methods 
which typically require thousands of BCUs.

\subsection{\textit{BUS}-based Recommendation}

\begin{figure}[H]
\begin{minipage}{1.0\linewidth}
  \includegraphics[width=\linewidth]{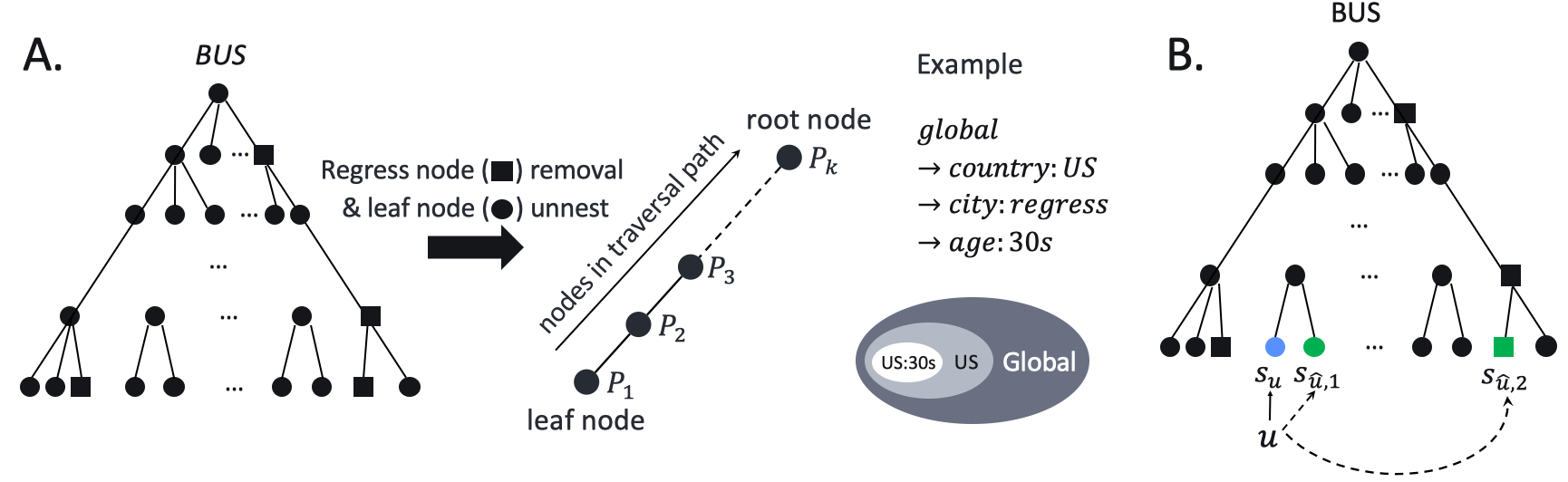}
  \captionof{figure}{\textit{BUS}-based recommendation. (A) A constructed \textit{BUS} tree undergoes \textit{regress} node removal and leaf node unnest. (B) Connection aware \textit{BUS}-based recommendation where content retrieved from user's own segment and connection segments.}
  \label{fig:recommendation}
\end{minipage}
\end{figure}

To generate retrieval candidates, each leaf node traverses upward to the first non-$regress$ ancestor, aggregating activity behaviors from all users under that ancestor, including descendants. The \textit{BUS} tree is transformed into $T'$ by collapsing regress attributes, and for each user, internal nodes are prioritized by distance to the leaf (Figure~\ref{fig:recommendation}A). Aggregated behaviors from these priority nodes are used individually or combined as the retrieval source for users assigned to the leaf. Nodes in \textit{BUS}-based recommendation are not mutually exclusive and a user contributes to multiple segments simultaneously. For example, the behavior of a user in $global \rightarrow US \rightarrow regress \rightarrow 30s$ segment contributes to $global$, $global \rightarrow US$, and $global \rightarrow US \rightarrow 30s$ nodes. This property distinguishes \textit{BUS}-based recommendation from standard clustering approaches, where users typically influence only their own cluster, while preserving interpretability and flexibility in candidate generation.

\vspace{-2em}

\subsection{Connection-aware \textit{BUS}-based Recommendation}

In product scenarios with a large number of content candidates, relying solely on a user’s own segment may limit personalization and raise fairness concerns. To mitigate this, we leverage users’ social graph to generate connection segments and combine content from both the user’s own segment ($s_u$) and the connected users’ segments ($s_{\hat u,1}$, $s_{\hat u,2}$,...) as retrieval sources (Figure~\ref{fig:recommendation}B). The social graph $G = (U, E)$ is constructed from the platform's friendship network, where $U$ represents all users and an edge $(u, u') \in E$ exists if users $u$ and $u'$ are mutual friends on the platform. For each user $u$, let $C_u = \{u' \in U \mid (u', u) \in E\}$ and $S_{\hat u} = \bigcup_{u' \in C_u} s_{u'}$ denote the connection segments. To reduce computation, we filter long-tailed segments using a percentile threshold $\phi$ (default 0.1) and round retained segment weights to 0.1. For example, connection distributions of $\{62\%,38\%\}$ and $\{61\%,39\%\}$ are both rounded to $\{0.6,0.4\}$, enabling shared computation. Aggregated behaviors from weighted connection segments are then combined with the user’s own segment to produce the final connection-aware \textit{BUS}-based recommendation, improving robustness, personalization, and fairness.

\section{Applications and Experiments}
\label{experiments}
\subsection{System Overview}

\setlength{\abovecaptionskip}{2pt}
\setlength{\belowcaptionskip}{2pt}
\renewcommand{\arraystretch}{1.0}
\vspace{-2em}
\begin{figure}[H]  
\begin{minipage}{1.0\linewidth}
    \centering
    \includegraphics[width=0.6\linewidth]{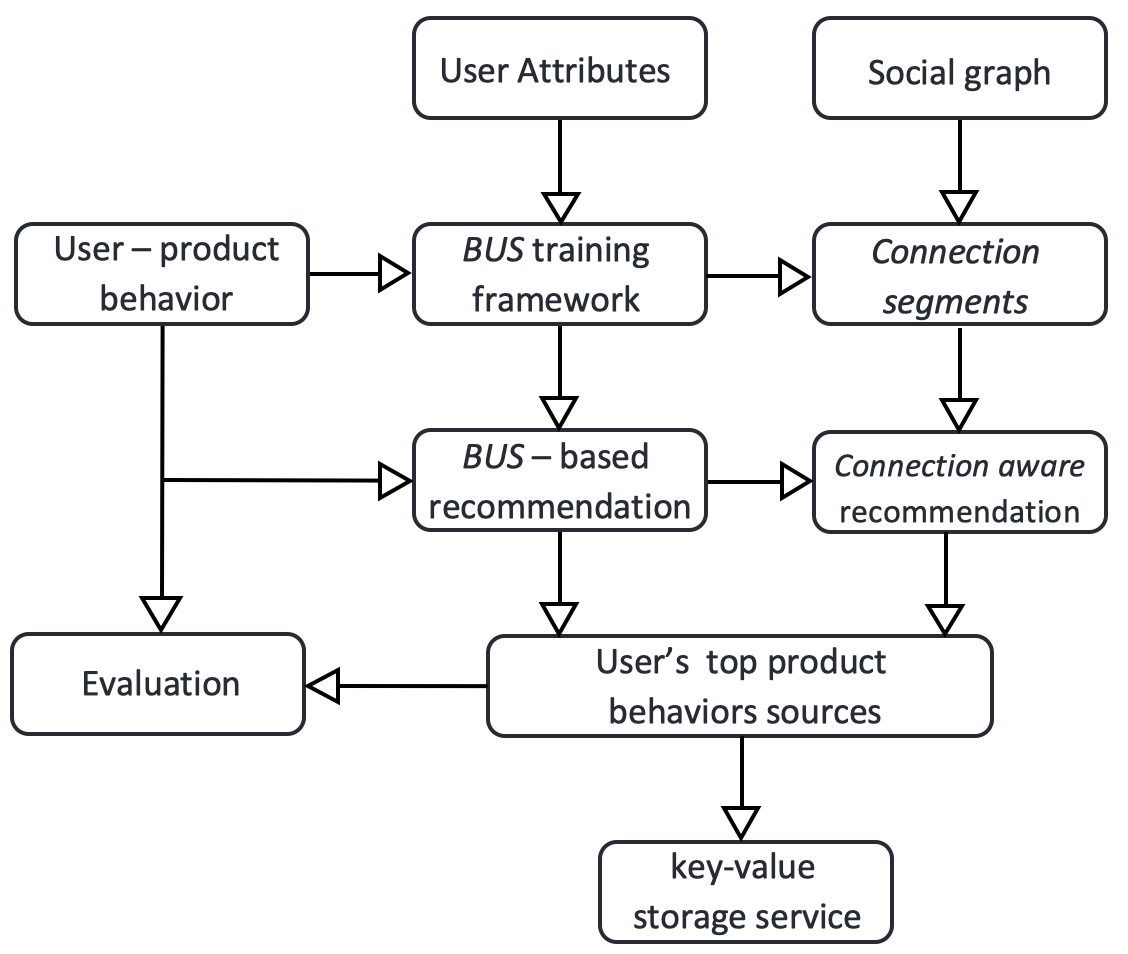}
    \caption{System overview of \textit{BUS}-based recommendation.}
    \label{fig:system}
\end{minipage}
\end{figure}
\vspace{-2em}

To support \textit{BUS} tree training and recommendation at billion-user scale, we implement the \textit{BUS} data structure in SQL, executed via a distributed query engine (Presto \cite{sethi2019presto}) within a workflow management service \cite{dataswarm}. We streamlined the \textit{BUS} training framework as a series of APIs to generate Dataswarm SQL codes that use the user attributes table and the user-product interaction behavior table to construct the \textit{BUS} tree iteratively in the workflow. We utilized relational database tables to model the tree node conceptual objects and optimized the SQL implementation by random sampling of up to 100K active users per staging segment in the computation of $\hat E_{AU_{s,p}}$ and storing the inherited reward to avoid duplicated computation.

The \textit{BUS} tree is periodically rebuilt using the latest user-product engagement data to maintain the long-term effectiveness of \textit{BUS}-based recommendations. For recommendation, top-K behaviors/contents are retrieved from each segment and optionally combined with connection segments. Popular behaviors are stored in Hive tables and cached in a distributed key-value store \cite{chen2016realtime} for fast real-time serving (Figure~\ref{fig:system}).

\subsection{Training \& Serving}

We evaluated \textit{BUS}-based recommendation in two personalization use cases: music ranking \cite{afchar2022explainability} and email notifications \cite{liu2022personalized}. In music ranking, millions of music and artist candidates were ranked to generate the top 100 IDs per \textit{BUS} segment, serving as one dedicated retrieval source. Prior work shows strong correlations between users' music habits and demographics \cite{cheng2014just, krismayer2019predicting}. For email notifications, we ranked 24-hour buckets per segment and use the segment's activity pattern to optimize delivery timing \cite{liu2022personalized, liu2020reinforcement}.

We defined active users as monthly active users on the target product (MAU), and marginal users as non-MAU but with activity on the product in the next 7 days. We set $K=100$ and $K=24$ for NDCG in music ranking and email notifications, respectively, with $\omega=1.0$ and 11 categorical features. Numerical demographics were bucketized, and missing features assigned a value of `\textit{NULL}'. The \textit{BUS} tree is rebuilt weekly to incorporate the latest behaviors. 

\setlength{\abovecaptionskip}{2pt}
\setlength{\belowcaptionskip}{2pt}
\renewcommand{\arraystretch}{0.9}
\vspace{-1em}
\begin{figure}[H]
\begin{minipage}{1.0\linewidth}
    \centering
    \includegraphics[width=0.8\linewidth]{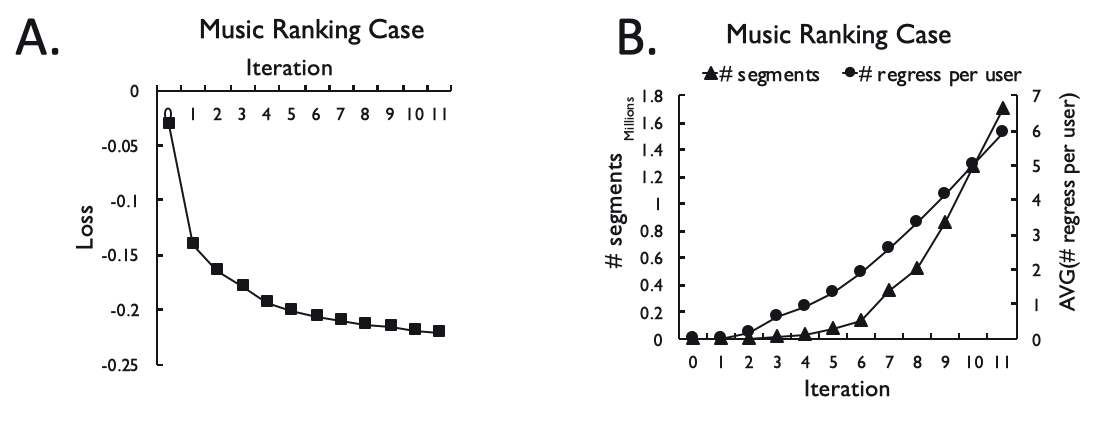}
    \captionof{figure}{Offline evaluation. (A) Overall loss in the \textit{BUS} tree construction. (B) The number of segments and \textit{regress} operators in the \textit{BUS} tree construction.}
    \label{fig:offline}
\end{minipage}
\end{figure}
\vspace{-1em}

Consistent with Lemma 1, overall loss decreases monotonically during tree growth (Figure~\ref{fig:offline}A), and $regress$ nodes are more frequent at lower levels (Figure~\ref{fig:offline}B), replacing ~50\% of attributes on average. Periodic rebuilding dynamically changes segment grouping, balancing stability and plasticity in recommendations \cite{fayyaz2020recommendation}. The \textit{BUS}-based recommendation sources are updated on a daily basis to provide the most recent segment's product behaviors or content candidates. In connection-aware recommendation, segment percentile threshold $\phi=0.1$ filters connection segments and popular candidates are ranked by a utility function:
\[
U_{u,c} = P_{c,s_u} + \sum_{s \in S_{\hat u}} w_{s,u} \cdot P_{c,s}
\]
where $P_{c,s}$ is candidate popularity in segment $s$, $s_u$ is 
the user's own segment, $S_{\hat u}$ denotes the user's connection segments, and $w_{s,u}$ is the weight of connection segment $s$ (e.g., the proportion of the user's friends belonging to that segment).

\subsection{Offline Evaluation}
Deep learning approaches such as neural collaborative filtering, graph neural networks, and embedding-based methods require sufficient interaction history to learn meaningful user representations. For marginal users with limited or no engagement data, these methods cannot generate reliable embeddings, making direct comparison infeasible. We focus on evaluating retrieval sources for marginal users, as our approach specifically targets this cold-start scenario where only categorical attributes are available. This complements rather than 
replaces deep learning methods for active users. Accordingly, our comparison is between \textit{BUS}-based recommendation and one-hot encoding-based aggregation.

\setlength{\abovecaptionskip}{2pt}
\setlength{\belowcaptionskip}{2pt}
\renewcommand{\arraystretch}{0.9}
\vspace{-2em}
\begin{figure}[H]
\begin{minipage}{1.0\linewidth}
  \includegraphics[width=\linewidth]{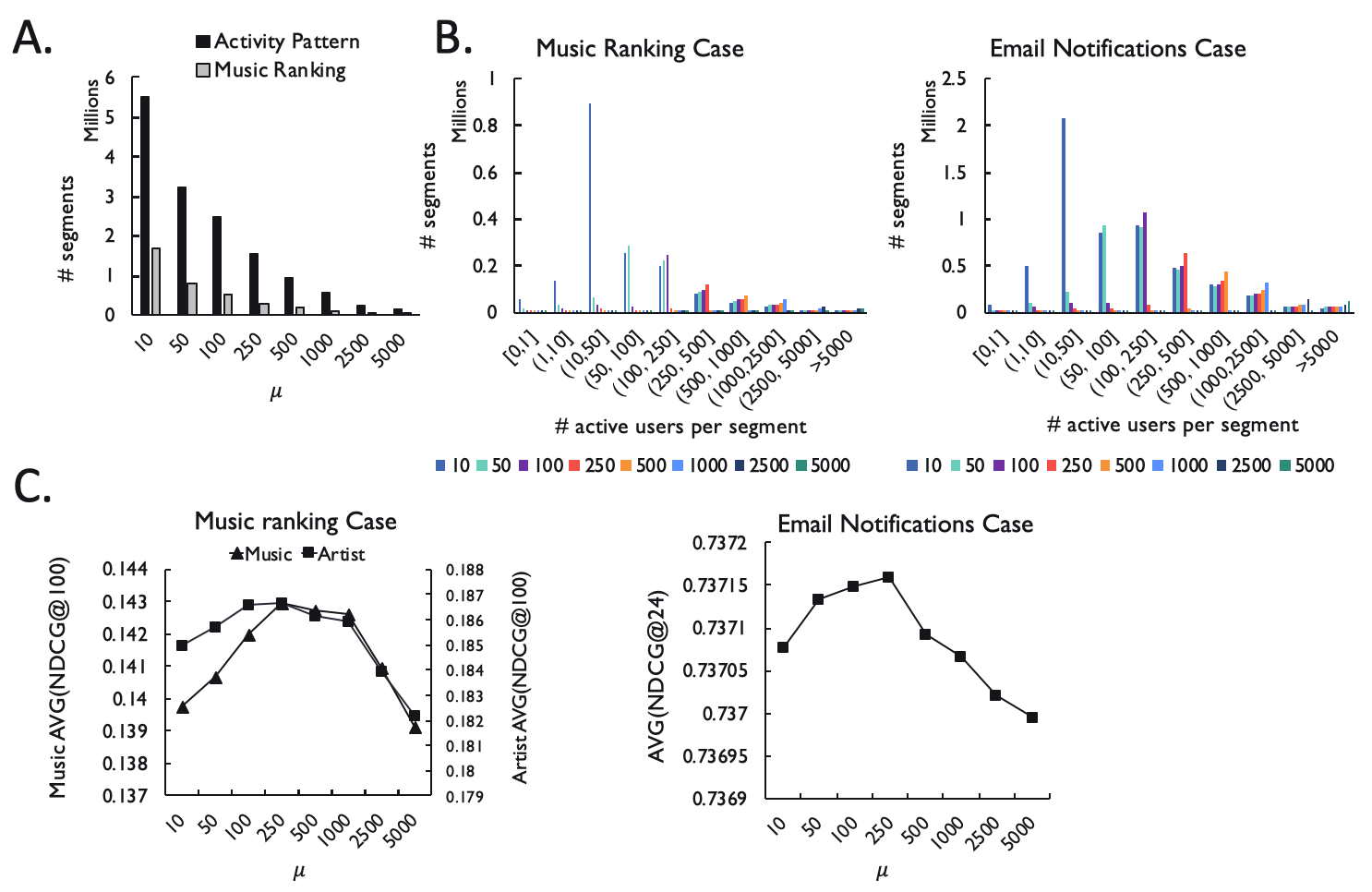}
  \captionof{figure}{Effect of segment size control in the \textit{BUS} tree construction. (A) Number of leaf node segments in \textit{BUS} trees built with varying minimum active user thresholds ($\mu$). (B) Size distribution of \textit{BUS} leaf node segments across different buckets for varying $\mu$ values. (C) NDCG scores for \textit{BUS} trees built with varying $\mu$.}
  \label{fig:size}
\end{minipage}
\end{figure}
\vspace{-2em}

We constructed 8 \textit{BUS} trees with minimum active user thresholds ($\mu$) from 10 to 5,000 and evaluated ranking quality by computing NDCG using marginal users' actual engagement behaviors in the subsequent 7 days as ground truth. Increasing $\mu$ reduced segment number and size (Figures~\ref{fig:size}A–B). Both use cases achieved the highest ranking quality at $\mu=250$ 
(Figure~\ref{fig:size}C), outperforming one-hot aggregation (Table~\ref{result-table}), suggesting that suitable segment size control mitigates overfitting. We also observed different product use cases exhibited distinct attribute ordering in \textit{BUS} trees, and $\mu$ influenced this ordering.

In the music ranking use case, we further implemented the connection-aware \textit{BUS}-based recommendation, generating 400-500 million distinct connection segments shared across billions of users. We observed that for 70\% of users, their own segment is not overlapped with their connection segments. This approach further boosted the music NDCG@100 from 0.1429 (Table~\ref{result-table}) to 0.150, with an additional computational cost of approximately 500--600 BCUs.

\setlength{\abovecaptionskip}{2pt}
\setlength{\belowcaptionskip}{2pt}
\renewcommand{\arraystretch}{0.9}
\vspace{-2em}
\begin{table}[H]
\centering
\caption{Ranking quality of \textit{BUS}-based vs. one-hot aggregation}
\label{result-table}
\begin{tabular}{lcccccc}
\toprule
Case & \multicolumn{3}{c}{Music Ranking} & \multicolumn{2}{c}{Email Notifications} \\
\cmidrule(lr){2-4} \cmidrule(lr){5-6}
Metric & music NDCG@100 | & artist NDCG@100 | & \#Segments | & NDCG@24 | & \#Segments \\
\midrule
\textit{BUS} & 0.1429 & 0.1867 & 303K & 0.7372 & 1.7M \\
City & 0.1104 & 0.1536 & 195K & 0.7355 & 195K \\
City + Age + Gender & 0.1325 & 0.1770 & 5M & 0.7365 & 5M \\
\bottomrule
\end{tabular}
\end{table}
\vspace{-2em}

\subsection{Online Experiments}
We conducted two 30-day online experiments to assess the impact of \textit{BUS}-based recommendation. The analysis is based on the aggregated results from the last 7 days of each experiment. In the email notifications, we evaluated an ensemble approach combining \textit{BUS}-based segment's activity patterns with the pointwise learning-based personalized prediction method \cite{liu2022personalized} (our baseline) to optimize email notification delivery times. We observed that the improved user activity pattern leads to statistically significantly ($p < 0.05$) increases in both email click-through rates and daily user app engagement (Table~\ref{online_combined}). Both marginal and active users became more active in the treatment, which suggests that the \textit{BUS} user segment can serve as a powerful collaborative filtering mechanism by capturing and disseminating cross-behavioral signals. 

\setlength{\abovecaptionskip}{2pt}
\setlength{\belowcaptionskip}{2pt}
\renewcommand{\arraystretch}{0.9}
\vspace{-2em}
\begin{table}[H]
\centering
\caption{Online experiments of \textit{BUS}-based recommendation.}
\label{online_combined}
\begin{tabular}{l|c|c|c}
\toprule
Experiment / Metric & Overall & Active & Marginal \\
\midrule
\textbf{Email notifications (\textit{BUS} vs. no-\textit{BUS})} & & & \\
Click-through rate & +2.95\% & +3.10\% & +2.36\% \\
Daily active users & +0.057\% & +0.021\% & +0.330\% \\
\midrule
\textbf{Music ranking (\textit{BUS} vs. no-\textit{BUS})} & & & \\
Daily active producers & +0.126\% & +0.11\% & +0.91\% \\
\midrule
\textbf{Music ranking (Connection-aware \textit{BUS} vs. \textit{BUS})} & & & \\
Daily active producers & +0.045\% & +0.043\% & +0.086\% \\
Shared rate & +0.16\% & +0.13\% & +0.58\% \\
\bottomrule
\end{tabular}
\end{table}
\vspace{-2em}

In music ranking, we evaluated the integration of \textit{BUS}-based recommendation as an additional retrieval source to provide the top 50 music IDs and artist IDs per segment. We compared this source with existing retrieval sources, which serve as our baselines and are based on various deep learning and collaborative filtering methods. Our results showed that \textit{BUS}-based recommendation leads to a statistically significant increase in content producers activities for both active and marginal producers (Table~\ref{online_combined}). Although we saw an increase in production, the content shared rate among users' friends was not significantly improved. To improve recommendation fairness and diversity, we introduced connection-aware \textit{BUS}-based recommendation and compared them with \textit{BUS}-based recommendation derived solely from the user’s own segment. This approach not only further boosts producer activity but also leads to a statistically significant increase in content sharing among friends, demonstrating a positive ecosystem impact.

\section{Conclusion}

We propose a hierarchical tree structure, \textit{BUS}, representing diverse user attributes to generate interpretable user segments and enhance attribute-based recommendations across multiple product scenarios. By combining list-wise learning-to-rank objectives with tree-based segmentation, \textit{BUS} produces semantically coherent user groups. Industrial-scale experiments in music ranking and email notifications show substantial improvements in key metrics, demonstrating efficiency and effectiveness. The \textit{BUS} framework is highly extensible and infrastructure-friendly, supporting diverse objectives and entity types beyond users. To our knowledge, this is the first large-scale deployment of a list-wise learning-to-rank framework leveraging tree-based recommendation with diverse entity attributes. Further studies are needed to clarify the optimal values of $\omega$ and $\phi$, and the strategy for combining \textit{BUS} with embedding methods for users transitioning from marginal to active status.

Looking forward, integrating LLMs with \textit{BUS} offers promising opportunities. Each segment provides an explicit, semantically consistent audience, enabling audience-level prompts for content summarization, creative generation, and insight extraction. Segment semantics can also be incorporated into ranking systems, bridging audience representation with content optimization. Thus, \textit{BUS} provides a unified foundation for recommendation, ranking, and generative AI applications.

\section{Acknowledgments}
We thank Yuankai Ge, Weizhe Shi, and partner teams for support and contributions.

\paragraph{Author Contributions}
Conceptualization and methodology: Y. Liu; experiments and analysis: Y. Liu, 
J. Kang, S. Iyer, I. Malik, R. Li, J. Wang, M. Liu, X. Lu, X. Zhao; supervision 
and administration: D. Wang, I. Liu, W. Shi, F. Liang, Y. Yu; manuscript 
preparation: Y. Liu. All authors reviewed and approved the final manuscript.

\vspace{1em}
\renewcommand{\bibname}{Reference}
\begingroup
\let\clearpage\relax
\let\cleardoublepage\relax
\bibliographystyle{splncs04}
\setlength{\parskip}{0pt}
\setlength{\itemsep}{0pt}
\setlength{\parsep}{0pt}
\setlength{\topsep}{0pt}

\endgroup
\end{document}